\documentclass[10pt]{article}
\usepackage[left=1.3in,top=1.0in,right=1.3in,bottom=1.0in]{geometry}
\usepackage{palatino}
\usepackage{graphicx}
\usepackage{subfigure}
\usepackage{amsfonts}
\usepackage{multirow}
\author{\begin{minipage}[\textwidth]{3in}\center Hamada A. Nayel \\  Department of Computer Science \\ 
             Benha University, Egypt\\\texttt{hamada.ali@fci.bu.edu.eg}\end{minipage}
\begin{minipage}[\textwidth]{3in}\center H. L. Shashirekha \\  Department of Computer Science\\ 
             Mangalore University, India\\\texttt{hlsrekha@gmail.com} \end{minipage}}
\title{Integrating Dictionary Feature into A Deep Learning Model for Disease Named Entity Recognition}
\date{}
\begin{document}
\maketitle
\begin{abstract}
{In recent years, Deep Learning (DL) models are becoming important due to their demonstrated success at overcoming complex learning problems. DL models have been applied effectively for different Natural Language Processing (NLP) tasks such as part-of-Speech (PoS) tagging and Machine Translation (MT). Disease Named Entity Recognition (Disease-NER) is a crucial task which aims at extracting disease Named Entities (NEs) from text. In this paper, a DL model for Disease-NER using dictionary information is proposed and evaluated on National Center for Biotechnology Information (NCBI) disease corpus and BC5CDR dataset. Word embeddings trained over general domain texts as well as biomedical texts have been used to represent input to the proposed model. This study also compares two different Segment Representation (SR) schemes, namely IOB2 and IOBES for Disease-NER. The results illustrate that using dictionary information, pre-trained word embeddings, character embeddings and CRF with global score improves the performance of Disease-NER system.}
\end{abstract}
\section{Introduction}
\begin{figure}[htb!]
\begin{center}
\includegraphics[height=1.6in, width=0.8\textwidth]{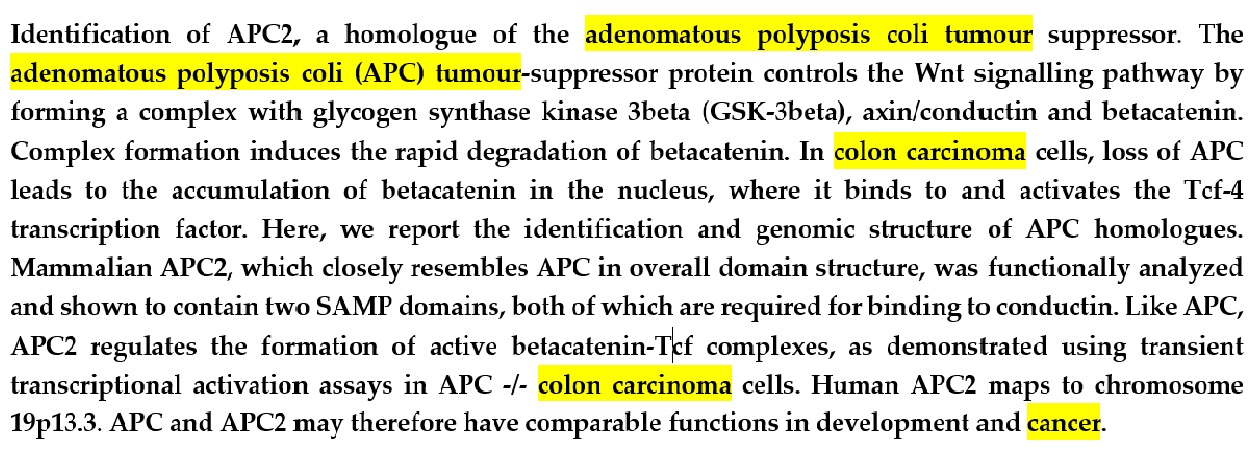}
\caption{\label{dexmp}An abstract with disease mentions highlighted}
\end{center}
\end{figure}

\begin{figure}[htb!]
\begin{center}
\includegraphics[height=2.1in, width=0.85\textwidth]{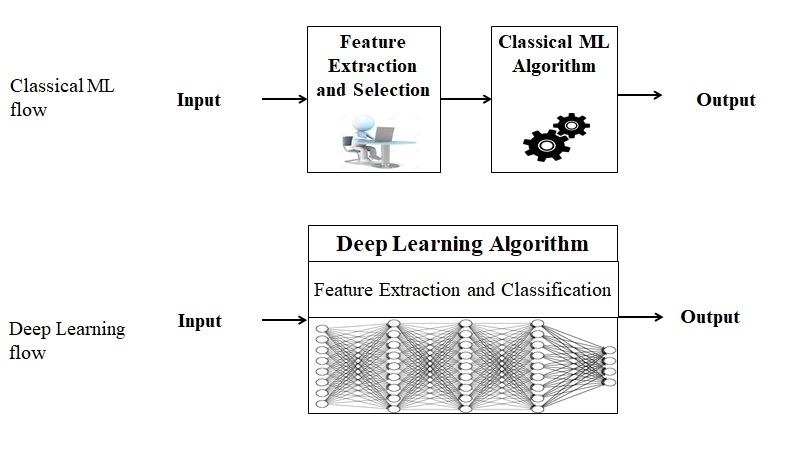}
\caption{\label{deep}Classical and deep learning flows}
\end{center}
\end{figure}

Disease is a principle Biomedical Named Entity (BioNE), which has got attention by biomedical research due to increase in research in health and the impact of disease on public life. Disease-NER is a challenging problem due to multiple challenges such as ambiguity (\emph{same word or phrase refers to different entities}), synonyms (\emph{an entity can be denoted by various names in a synonym relation}), multi-word NEs (most of disease NEs consist of multiple words) and nested NEs (\emph{one NE may occur within a longer NE}). Further, abbreviations which are used frequently in biomedical literature are the main sources of ambiguity. For example, ``\emph{AS}" may refer to ``\emph{Asperger Syndrome}" or ``\emph{Autism Spectrum}" or ``\emph{Aortic Stenosis}" or ``\emph{Ankylosing Spondylitis}" as well as ``\emph{Angleman Syndrome}". In such cases to which entity an abbreviation refers to has to be resolved depending on the context. Figure \ref{dexmp} shows an example of an abstract with disease mentions highlighted.
Different approaches have been used for Disease-NER, such as dictionary-based approach, rule-based approach, Machine Learning (ML) approach and hybrid approach \cite{NayelThesis}\cite{nayel-shashirekha:2017:W17-75}. Deep Learning (DL) is a big trend in ML, which promises powerful and fast ML algorithms moving closer to the performance of Artificial Intelligence (AI) systems. It is about learning multiple levels of representation and abstraction that help to make sense of any data such as images, sound, and text. Automatically learning features at multiple levels of abstraction allow a system to learn complex functions mapping the input to the output directly from data, without depending on human-crafted features. DL employs the multi-layer Artificial Neural Networks (ANN) for increasingly richer functionality. While the concept of classical ML is characterized as learning a model to make predictions based on past observations, DL approaches are characterized by learning to not only predict but also to correctly represent the data such that it is suitable for prediction. Figure \ref{deep} shows the difference between classical ML flow and DL flow.\\
\\Given a large set of desired input-output mapping, DL approaches work by feeding the data into an ANN that produces consecutive transformations of the input until a final transformation predicts the output. These transformations are learnt from the given input-output mappings, such that each transformation makes it easier to relate the data to the desired label.
\\Of late, DL algorithms based on ANNs \cite{Li2017,Xu2018} are being used to a larger extent for various NLP tasks such as Biomedical Named Entity Recognition (BioNER) \cite{hamadaYuji} relation extraction for biomedical texts \cite{relationANNLi20171} and biomedical event extraction \cite{rao-EtAl:2017:BioNLP17}. In this paper, an efficient DL model using ANN for Disease-NER has been developed. NCBI and BC5CDR datasets are used for evaluation of the proposed model and the results are reported in terms of f1-measure.
\section{Background}
\subsection{Artificial Neural Networks (ANN)}
\begin{figure}[htb!]
\centering
\subfigure[\label{s1}Structure of a biological neuron]{\includegraphics[height=1.5in,width=0.49\textwidth] {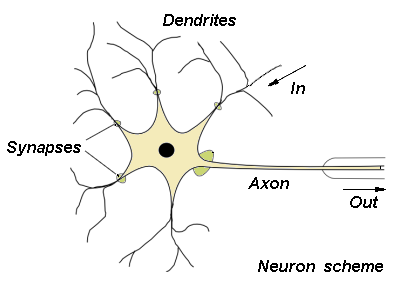}}
\subfigure[\label{s2}A simple neuron model]{\includegraphics[height=1.5in,width=0.49\textwidth] {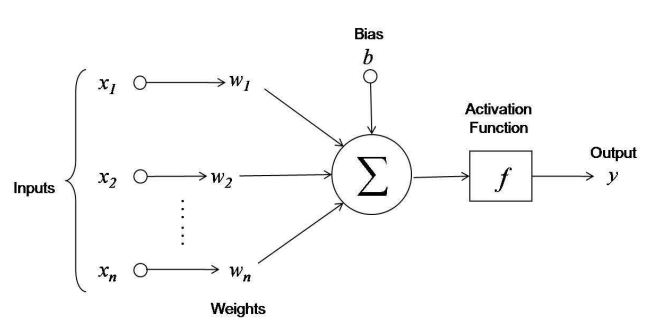}}
\caption{Structure of a biological neuron and a simple neuron model\label{neuron}}
\end{figure}

\begin{figure}[h!]
\begin{center}
\includegraphics[height=1.5in, width=0.80\textwidth]{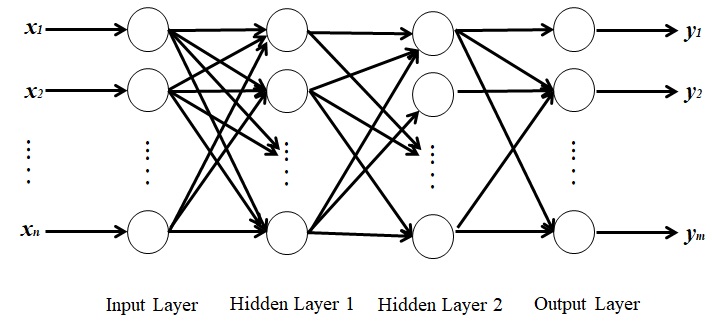}
\caption{\label{annModel}Structure of a simple feed-forward ANN}
\end{center}
\end{figure}

\begin{figure}[htb!]
\begin{center}
\includegraphics[height=1.5in, width=0.50\textwidth]{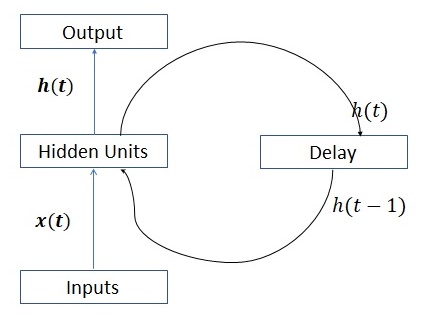}
\caption{\label{rnn}A Simple RNN Structure}
\end{center}
\end{figure}

\begin{figure}[htb!]
\begin{center}
\includegraphics[height=1.7in, width=0.60\textwidth]{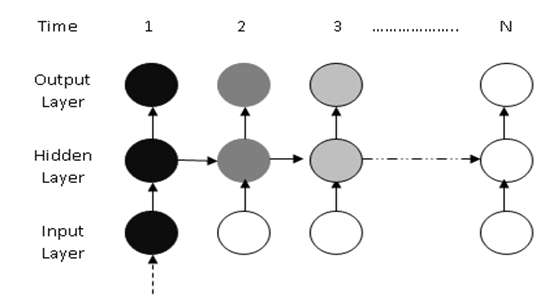}
\caption{\label{vanish}Gradient vanishing problem for RNN}
\end{center}
\end{figure}

ANN are inspired by the mechanism of brain computation which consists of computational units called neurons. However, connections between ANN and the brain are in fact rather slim. In the metaphor, a neuron has scalar inputs with associated weights and outputs. The neuron multiplies each input by its weight, sums them and transforms to a working output through applying a non linear function called activation function. The structure of a biological neuron and an artificial neuron model with $n$ inputs and one output is shown in Figure \ref{neuron}. In this example, a neuron receives simultaneous inputs $X=(x_1,x_2,\ldots,x_n)$ associated with weights $W=(w_1,w_2,\ldots,w_n)$, a bias $b$ and calculates the output as $y = f(W\cdot X + b)$ where $f$ is the activation function.
%\begin{equation}\label{basicann}
%  y = f(W\cdot X + b)
%\end{equation}
ANN comprises of a large number of neurons within different layers with each layer having a specific task. An ANN model basically consists of three layers: an input layer, one or more hidden layers and an output layer. Input layer contains a set of neurons called input nodes, which receive raw inputs directly. The hidden layers receive the data from the input nodes and responsible for processing these data by calculating the weights of neurons at each layer. These weights are called connection weights and passed from one node to another. Number of nodes in hidden layers influences the number of connections as well as computational complexity. During learning connection weights are adjusted to be able to predict the correct class label of the input. Using multiple hidden layers helps in detecting more features while learning the model. Output layer receives the processed data and uses its activation function to generate final output. This kind of ANN where information flows in one direction from input layer to output layer through one or more hidden layers is called feed-forward ANN. Figure \ref{annModel} shows an example of a feed-forward ANN with two hidden layers. An ANN is called fully connected if each node in a layer is connected to all nodes in the subsequent layer.
\newline Recurrent Neural Networks (RNN) is a type of ANN in which hidden layer neurons has self-connections which means output depends not only on the present inputs but also on the previous neuron state. A simple form of RNN which contains an ANN with the previous set of hidden unit activations feeding back into the network along with the inputs is shown in Figure \ref{rnn}. The activations are updated at each time step $t$ and a delay unit has been introduced to hold activations until they are processed at the next time step. The input vector $x_0$ at time stamp $t=0$ processed using RNN structure is as follows:
\begin{equation}\label{rnneq}
   h_t = f_W(h_{t-1},x_t)
\end{equation}
where, $h_t$ is the output at time stamp $t$, $h_{t-1}$ is the output at time stamp $t-1$, $f_W$ is an activation function with parameter $W$ and $x_t$ is the input vector at the time stamp $t$.\\
In case of long sequences, RNNs are biased towards their most recent inputs in the sequence due to the gradient vanishing problem \cite{Bengio:1994:LLD:2325857.2328340,Pascanu:2013:DTR:3042817.3043083}. While calculating weights in RNN for each time stamp $t$, the gradients of error get smaller and smaller as moving backward in the network and gradually vanish. Thus, the neuron in the earlier layers learns very slowly as compared to the neurons in the later layers. Earlier layers in the network are important as they are responsible to learn and detect patterns and are the building blocks of the RNN. Figure \ref{vanish} shows an example of gradient vanishing problem. In this figure, the dark shade of the node indicates the sensitivity over time of the network nodes to the input at first time stamp. The sensitivity decreases exponentially over time as new inputs overwrite the activation of hidden unit and the network forgets the input at first time stamp. To overcome the gradient vanishing problem, S. Hochreiter and  J. Schmidhuber \cite{LSTM} introduced a new RNN architecture called Long Short-Term Memory (LSTM).
\subsubsection{Long Short-Term Memory}
\begin{figure}[htb!]
\begin{center}
\includegraphics[height=1.7in, width=0.40\textwidth]{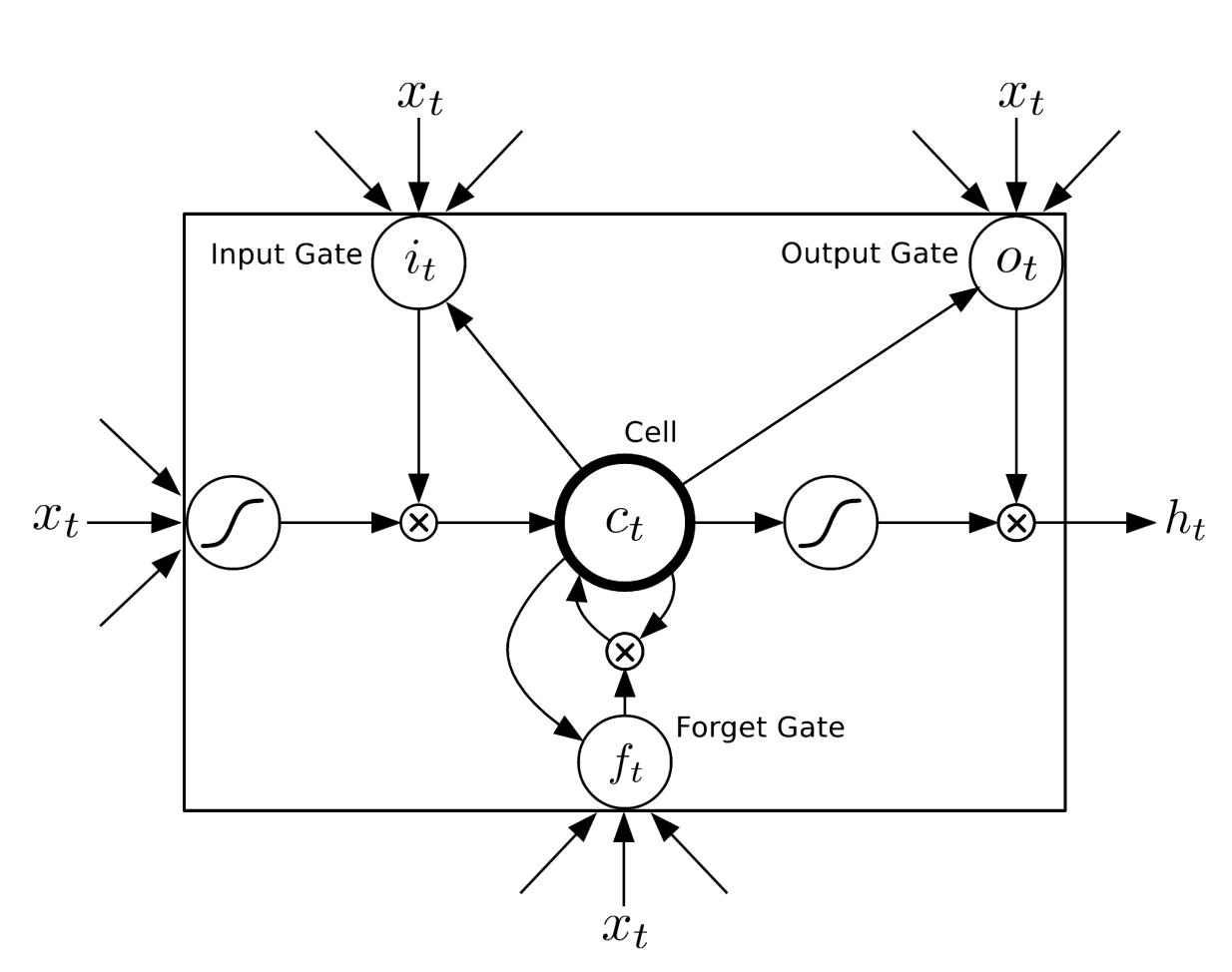}
\caption{\label{lstm4}Structure of LSTM unit}
\end{center}
\end{figure}

\begin{figure}[h!]
\begin{center}
\includegraphics[height=1.9in, width=0.60\textwidth]{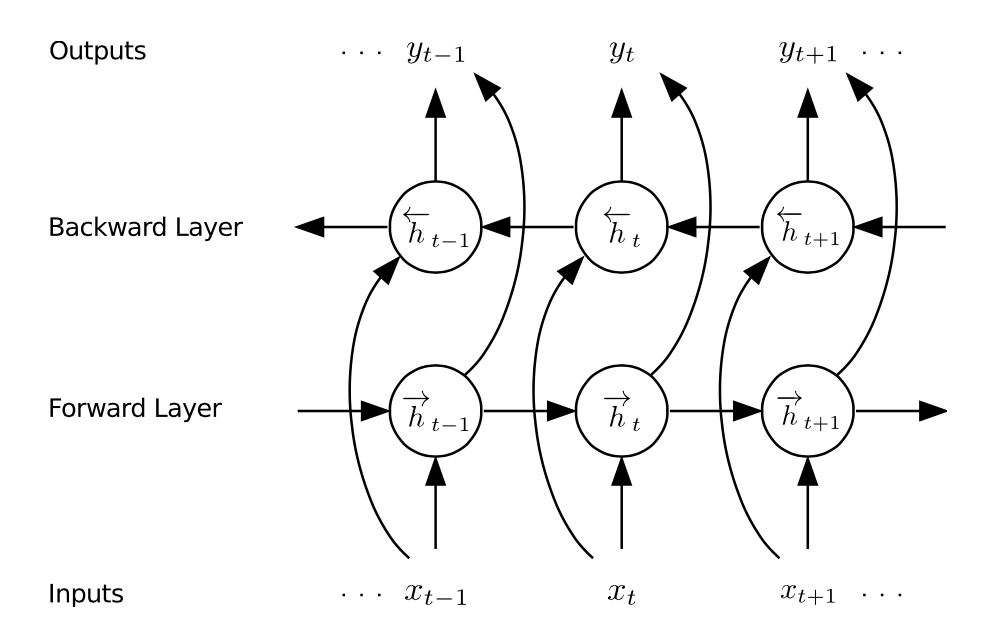}
\caption{\label{bilstm}Structure of Bidirectional LSTM}
\end{center}
\end{figure}

LSTM \cite{LSTM} is a kind of RNN which handles sequences of arbitrary length and is able to model dependencies between far apart sequence elements as well as consecutive elements. The LSTM architecture consists of a set of RNNs known as memory blocks. Each block contains self-connected memory cell ($m_t$) and three multiplicative units namely input ($i_t$), output ($o_t$) and forget ($f_t$) gates, that provide continuous peers of write, read and reset operations for the cells as shown in Figure \ref{lstm4}.
These gates regulate the information in memory cell and consists of a sigmoid function. The input gate regulates the proportion of history information that will be kept in memory cell and the output gate regulates the proportion of information stored in the memory cell which will influence other neurons. Forget gate can modify the memory cell by allowing the cell either to remember or forget its previous state. The complete details of LSTM architecture is described by S. Hochreiter and J. Schmidhuber \cite{LSTM}.\\
\\In the sequence labeling problem, the typical input is a sequence of input vectors and the output is a sequence of output tags. NER can be considered as a sequence labeling problem where the sentence $X$ consisting of words $(w_1, w_2,...,w_n)$ is given as input and the required output is the sequence of tags $T=(t_1,t_2,..., t_n)$ that represents the class labels of the words. LSTM is suitable to apply for NER as it can remember up to the first word in the sentence. However, one shortcoming of LSTM is that they process the input only in left context. But, in NER it is beneficial to have access to both left and right contexts as the output tag of a word depends on few previous and few next words (context window). This problem is overcome by a Bidirectional LSTM (BiLSTM) \cite{GRAVES2005602} where each sequence is presented in forward and backward direction to two separate hidden states to capture left and right context information respectively. Then the outputs of two hidden states are concatenated to form the final output as shown in Figure \ref{bilstm}.
\\To apply BiLSTM for NER, a sentence $X=(w_1, w_2,...,w_n)$ has to be fed to a BiLSTM and each word has to be replaced by its vector representation. Applying BiLSTM structure on these representations outputs a vector $Y=(t_1, t_2,...,t_n)$ representing the corresponding tags.\\
\\ANN in general can be applied for language understanding and NLP tasks such as speech recognition, MT and text summarization using the concept of language modeling (LM). LM is a probabilistic model that is able to predict the next word in the sequence given the words that precede it. The central component in ANN language modelling is the use of an embedding layer which maps discrete symbols (words, characters) to continuous vectors in a relatively low-dimensional space. Vector representation of words and documents has been a leading approach in information retrieval and computational semantics \cite{Turney:2010}. The primitive method of representing a document as a vector is the bag-of-words model. If $m$ is the size of the vocabulary, a document is assigned a $m$-dimensional vector with non-zero entries for words corresponding to the entries occurred in the document and zero entries for rest of the words. These are used to learn $n$-dimensional real-valued vector representations of words in $\mathbb{R}^n$. After the rise in popularity of ANNs, a new approach for representing lexical semantics has become the new default for novel models.  These real-valued vectors are called "word embeddings" and have become a primary part of models for various applications, such as information retrieval \cite{Ganguly:2015:WEB:2766462.2767780}, sentiment analysis \cite{dos2014deep}, automatic summarization \cite{YOUSEFIAZAR201793}and question answering \cite{sukhbaatar2015end}.
\subsection{Word Embeddings}
\begin{figure}[htp!]
\begin{center}
\includegraphics[height=1.5in, width=0.35\textwidth]{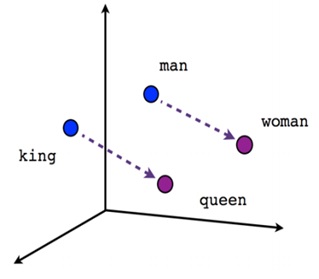}
\caption{\label{wordvector}Example of word vector calculations}
\end{center}
\end{figure}

Word embeddings is a distributed representation of words in a vector space that captures semantic and syntactic information for words \cite{mikolov2013distributed}. The basic idea behind word embeddings is to use distributional similarity based representations by representing a word by means of its neighbors. Distributed representations of words help to enhance the performance of learning algorithms in various NLP tasks by grouping similar words. Mikolov et al. \cite{mikolov2013efficient} introduced the skip-gram model and Continuous Bag of Words (CBOW) model for learning word embeddings from huge amounts of text data. An ANN structure has been used for learning word embeddings which encode many linguistics regularities and patterns explicitly. Some of these patterns can be represented as linear operations, e.g. the result of vector calculation: $\overrightarrow{king} - \overrightarrow{man} + \overrightarrow{woman}$ is closer to $\overrightarrow{queen}$ than to any other word vector as shown in Figure \ref{wordvector}.
CBOW model trains the word embeddings by predicting a word in a sentence using its surrounding words, while Skip-gram model trains the word embeddings by predicting the surrounding words for a given word in the input layer. The word embeddings $e_t$ representing a word located at the \emph{i}-th position in a sentence is calculated by maximizing the average log probability as follows:-
\begin{itemize}
  \item In CBOW model
  \[\frac{1}{T}\sum_{t=1}^{T}\log p(e_t|e_{t-\frac{n}{2}},\dots,e_{t-1},e_{t+1},\dots,e_{t+\frac{n}{2}})\]
  \item In Skip-gram model
  \[\frac{1}{T}\sum_{t=1}^{T}\log p(e_{t-\frac{n}{2}},\dots,e_{t-1},e_{t+1},\dots,e_{t+\frac{n}{2}}|e_t)\]
\end{itemize}
where $e_{t-\frac{n}{2}},\dots,e_{t-1}$ are vectors for $\frac{n}{2}$ preceding words, $e_{t+1},\dots,e_{t+\frac{n}{2}}$ are vectors for $\frac{n}{2}$ subsequent words, and $T$ is the number of tokens in the sentence.
\paragraph*{\textbf{Character-level Embeddings}} Word embeddings maintain semantic and syntactic information of words, but does not capture orthographical information at character level such as capitalization, numeric characters, special characters and hyphenation. However, such information plays an important role in Disease-NER as disease names can be characterized by the existence of combination of these information. Character-level word representation has been introduced to represent orthographical and morphological information \cite{DosSantos:2014:LCR:3044805.3045095}.
\section{Related Works}
Researchers have explored many approaches for Disease-NER. Robert Leaman et al. \cite{leamanDnorm} have addressed Disease-NER by learning the similarities between the disease mentions and concept names. They evaluated their approach using NCBI dataset and achieved a f1-measure of 80.9\%. This was the first work to use pairwise learning to rank Disease NER approach but the performance was not high. A multi-task learning approach applied to BioNER using Neural Network architecture by Gamal C. et al. \cite{Crichton2017} has achieved a f1-measure of 80.73\% for Disease-NER using NCBI dataset. In addition to NCBI dataset, 15 biomedical corpora have been used to train the system. These corpora have joint articles with NCBI test set and this affects the model evaluation. Leaman and Lu \cite{Leaman09062016} used semi-Markov models to build TaggerOne, a tool for BioNER which reported a high competitive f1-measure of 82.9\% for NCBI dataset. Sunil and Ashish \cite{sahu-anand:2016:P16-1} designed a RNN model for Disease-NER using Convolution Neural Network for representing character embeddings and bidirectional LSTM for word embeddings. A pre-trained word embeddings trained over a corpus of PubMed articles have been used for training the model and it is not enough to represent words in general domain. They evaluated their system on NCBI dataset and achieved f1-measure of 79.13\%. Wei et al. \cite{doi:10.1093/database/baw140} developed an ensemble-based system for Disease-NER. At the base level, they built CRF with a rule-based post-processing system and Bi-RNN based system. At the top level, they used SVM classifier for combining the results of the base systems. The proposed system uses manually extracted features for SVM and CRF as well as hand-crafted rules which depends on human experts. BC5CDR corpus has been used for evaluation and the model reported a f1-measure of 78.04\%.\\
\\BANNER \cite{leaman2008banner}, a tool which implements CRF algorithm for BioNER using general features such as orthographical, linguistic and syntactic dependency features has reported a f1-measure of 81.8\% on NCBI dataset. Xu et al. \cite{doi:10.1093/database/baw036} designed a system (CD-REST) for chemical-induced disease relations from biomedical texts. A CRF-based module for NER as first step is designed using character level, word level features, context features and distributed word representation features learned from external un-annotated corpus. They evaluated the system using BC5CDR dataset and reported a f1-measure of 84.43\%. Zhao et al. \cite{zhao2017disease} developed a system for Disease-NER based on convolutional neural network. The proposed system integrated with dictionary information and a post-processing module has been used for performance enhancements. NCBI and BC5CDR datasets have been used for evaluation of the proposed system and reported a f1-measure of 85.17\% and 87.83\% respectively. Haodi Li et al. \cite{doi:10.1093/database/baw077} designed an end-to-end system used for chemical-disease relation extraction. The proposed system uses an ensemble approach using SVM and CRF as base classifiers with SVM as a meta-classifier to build a module for Disease-NER. BC5CDR dataset has been used to evaluate the system and reported a f1-measure of 86.93\%. Hsin-Chun Lee et al. \cite{lee2015enhanced} presented an enhanced CRF-based system for Disease-NER. Rich feature set in addition to dictionary based features extracted from different lexicons have been used to train CRF. BC5CDR dataset has been used to evaluate has been used for system evaluation and reported f1-measure of 86.46\%. 
\paragraph*{} In this paper, a DL model has been designed for Disease-NER using BiLSTM for learning the model and CRF for decoding the results with the following objectives:
\begin{itemize}
	\item Using dictionary information for each token
	\item Using pre-trained word embeddings trained over a huge corpus of texts from biomedical domain as well as generic domain
	\item Learning a BiLSTM model for character-level word representations instead of using hand engineered features
\end{itemize}
\section{Methodology}
\begin{figure}[htp!]
\begin{center}
\includegraphics[height=2.0in, width=0.75\textwidth]{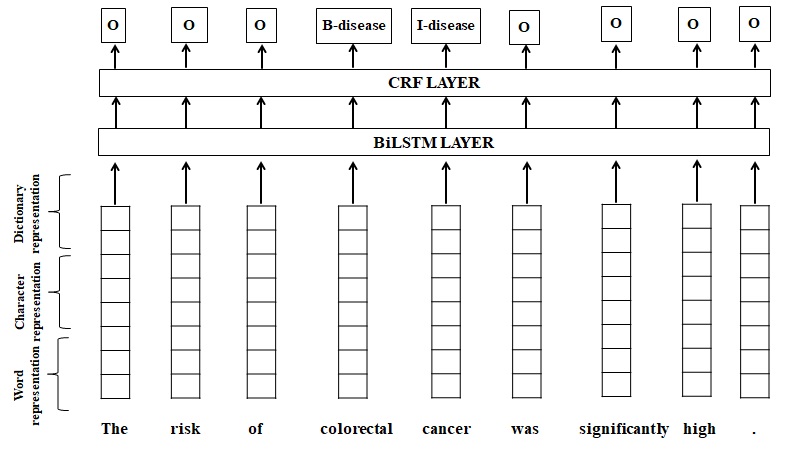}
\caption{\label{lstmcrf}General structure of the proposed model}
\end{center}
\end{figure}

\begin{figure}[!h]
\begin{center}
\includegraphics[height=2.0in ,width = 0.75\textwidth]{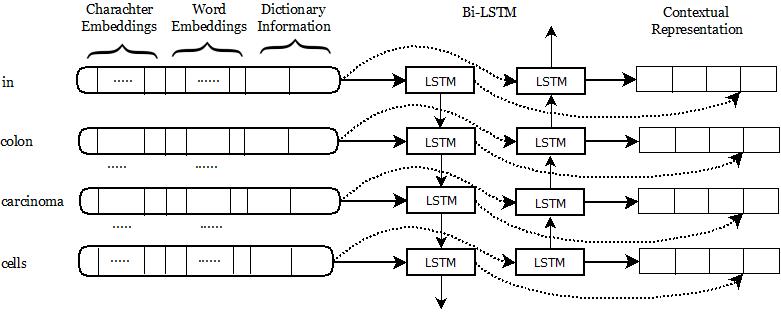}
\caption{\label{wrd}The contextual representation learning model}
\end{center}
\end{figure}

General structure of the proposed model is given in Figure \ref{lstmcrf}. Proposed model accepts a sequence of words, for example, the sentence ``\emph{The risk of colorectal cancer was significantly high.}" and the associated tags ``(O, O, O, B-Disease, I-Disease, O, O, O, O)" as input and gives a vector representation for each word containing information about the word itself and the neighbouring words within a sentence, denoted as contextual representation. In addition to word embeddings and character-level embeddings, dictionary information for each token has been extracted from MErged DIsease voCabulary (\textbf{MEDIC}) \cite{doi:10.1093/database/bar065}. MEDIC is a comprehensive and publicly available dictionary for disease entities which provides information including disease names, concept identifiers, definitions of diseases and synonyms. Dictionary information for each word is represented as a binary vector containing information about the existence of the token in the dictionary either solo or as a part of multi-word disease name or abbreviation or a synonym of a disease.
Character level representation is concatenated with word embeddings and the dictionary information. At this level, every word is represented as a vector comprising of character level information, word level information and dictionary information. Feeding the vector representation of word sequence of a sentence in direct and reverse order to a BiLSTM network will output a contextual representation for each word as shown in Figure \ref{wrd}.
\subsection{\label{decoding}Decoding}
\begin{figure}[htb!]
\centering
\subfigure[\label{decod1}Path of the sequence ((O, B-Disease, B-Disease, O)]{\includegraphics[height=1.2in,width=0.46\textwidth] {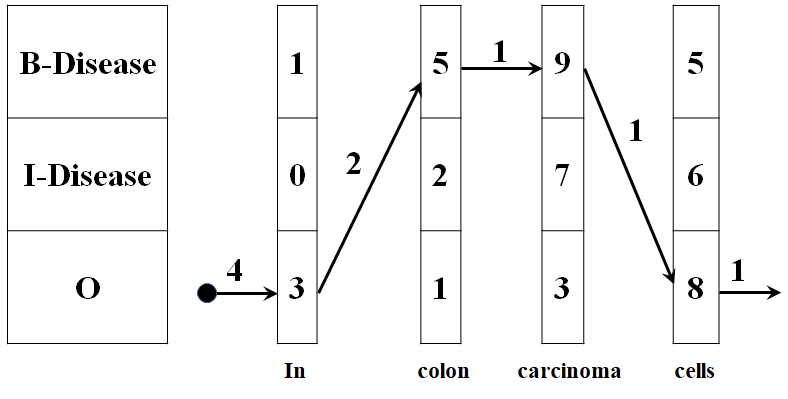}}
\subfigure[\label{decode2}Path of the sequence ((O, B-Disease, I-Disease, O)]{\includegraphics[height=1.2in,width=0.46\textwidth] {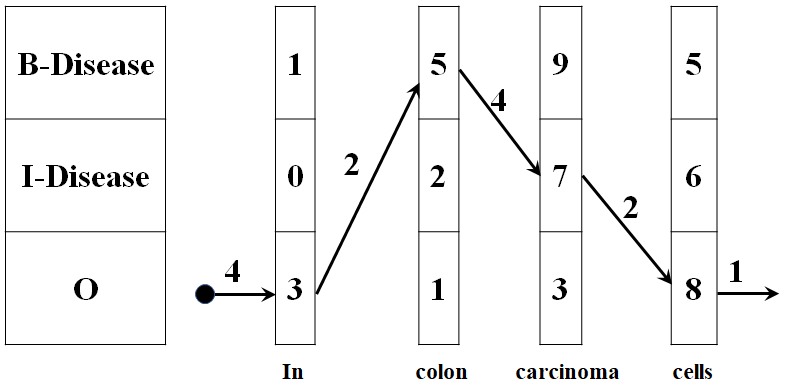}}
\caption{An example of decoding a text fragement\label{decode}}
\end{figure}

Decoding is the final step, which converts the contextual representations of the tokens into corresponding tags. For sequence labeling problem, it is beneficial to consider the correlations between labels of the tokens in the neighbourhoods and jointly decode the best chain of labels for a given input sentence. CRF \cite{Lafferty:2001:CRF:645530.655813} model is used for decoding as CRF considers the contextual information for decoding the label for each token. There are two approaches for calculating the scores of output tags; local scores and global scores. Local scores use scores represented in the final contextual representations for each word, while global scores use transition scores as well as scores represented in the final contextual representations for each word.\\
\\A fully connected neural network has been used to convert the contextual representations of the tokens to a vector where each entry corresponds to a score for each output tag. For an input sentence, $\textbf{X}=(x_1,x_2,...,x_m)$, $\textbf{Y}=(y_1,y_2,...,y_m)$ is the sequence of output tags, $S\in R^{T\times m}$ is the score matrix, where $T$ is number of predefined tags and $s_{i,j}\in S$ is the score of \emph{i}\textsuperscript{th} tag for the \emph{j}\textsuperscript{th} word. A global score, $Score\in R$ of sequence of tags $\textbf{Y}=(y_1,y_2,...,y_m)$ is defined as:
\[Score(y_1,y_2,...,y_m) = \sum_{t=1}^{m}\Big( s_{y_t,t} + Tr[y_t,y_{t+1}] \Big)\]
where, $Tr[y_t,y_{t+1}]$ is the score of assigning the tag $y_{t+1}$ given the tag $y_t$.\\
\noindent The sum of scores of all possible sequences of the tags for an input sentence $\textbf{X}$ is calculated as:
\[Z=\sum_{y'_i\in Y(x), 1\leq i\leq m}e^{Score(y'_1,y'_2,...,y'_m)}\]
\noindent Then given the sentence $\textbf{X}=(x_1,x_2,...,x_m)$, the conditional probability of a label sequence
$\textbf{Y}=(y_1,y_2,...,y_m)$ is defined as:
\[P(\textbf{Y}|\textbf{X}) = \frac{e^{Score(y_1,y_2,...,y_m)}}{Z} \]
The predicted tag sequence $\hat{y}\in Y(\textbf{x})$ is calculated as:
\[\hat{y}= \textbf{argmax} \{ P(\textbf{Y}|\textbf{X})\}\]
Example of decoding a contextual representation for text fragment ``\emph{In colon carcinoma cells}" is given in Figure \ref{decode}. In this example, numbers in columns are scores of assigning corresponding tags to the word. CRF is used to calculate the score of all paths and then the path with maximum score will be chosen. The global scores of two sequences are calculated as follows:\\

{\indent\indent{\itshape Score}(O, B-Disease, I-Disease, O)  = 4+3+2+5+4+7+2+8+1=36\\}
{\indent\indent{\itshape Score}(O, B-Disease, B-Disease, O) = 4+3+2+5+1+9+1+8+1=34\\}
\noindent The first path having higher score will be selected by CRF.
\\In addition to this approach, a local score can be used to decode the output vectors to tags as follows:
\[local\_score(y_1,y_2,...,y_m) = \sum_{t=1}^{m}s_{y_t,t}\]
The predicted tag sequence $\hat{y}\in Y(\textbf{x})$ is calculated as:
\[\hat{y}= \textbf{argmax} \{ local\_score(y_1,y_2,...,y_m)\}\]
\section{Datasets}
\subsection{\label{ncbidataset} NCBI Dataset}
NCBI disease corpus was introduced for disease name recognition and normalization \cite{DOGAN20141}. It is the most inclusive publicly available dataset annotated with disease mentions. The corpus was manually annotated by 14 medical practitioners. General statistics of the dataset is given Table \ref{sta}.
\begin{table}[!h]
\renewcommand{\arraystretch}{1.3}
\begin{center}
\begin{tabular}{l|rrrr}\hline
                                  & \textbf{Training}& \textbf{Dev}& \textbf{Test}& \textbf{Total}\\ \hline
\textbf{No. of Abstracts}         & 593                  & 100                     & 100              & 793           \\
\textbf{No. of Sentences}         & 5661                 & 939                     & 961              & 7261          \\
\textbf{Total Disease mentions}   & 5145                 & 787                     & 960              & 6892          \\
\textbf{Unique Disease mentions}  & 1710                 & 368                     & 427              & 2136          \\ \hline
\end{tabular}
\end{center}
\caption{\label{sta}Statistics of NCBI dataset}
\end{table}
\subsection{BC5CDR dataset}
BC5CDR dataset was created for BioCreative V Chemical Disease Relation (CDR) task and consists of 1500 PubMed articles with 5818 disease mentions \cite{bc5cdr}. The corpus was randomly split into three subsets: 500 each for training, testing and development sets. A BioNE class label named \texttt{DISEASE} and \texttt{O} (for non-BioNEs) are used to annotate the dataset.
\section{Experiments}
\subsection{Pre-processing}
\begin{figure}[!h]
\begin{center}
\includegraphics[height=2.1in,width = 0.75\textwidth]{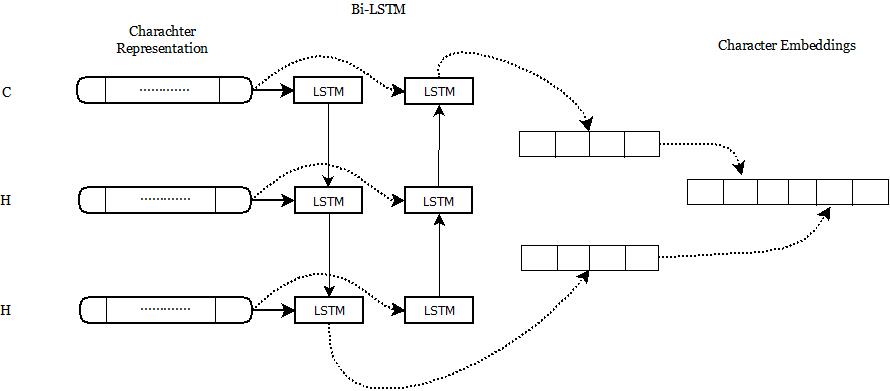}
\caption{Character representation learning model}
\label{chr}
\end{center}
\end{figure}

\indent Pre-trained word embeddings using Skip-gram model are used to represent tokens in the data set. This word embeddings\footnote{http://bio.nlplab.org/} model combines domain-specific texts (PMC and PubMed texts) with generic ones (English Wikipedia dump) for better representation of tokens. We have used pre-trained word embeddings trained by Sampo Pyysalo et al. \cite{moen2013distributional}, as constructing word embeddings requires a huge corpus and machines with high specifications. Using skip-gram model for training word embeddings improves the semantic and syntactic representation. Character-level embeddings are created by initializing vector representations for every character in the corpus and then the character representation corresponding to every character in a word are given in direct and reverse order to BiLSTM as shown in Figure \ref{chr}.
All numbers in the input are replaced with the value \texttt{NUM} (number), all letters are converted to lowercase and words that are not represented in the pre-trained word embeddings are marked as \texttt{UNK} (unknown). As the pre-trained word embeddings are of large size, a look-up table containing the word embeddings of all words in the dataset are extracted from the pre-trained word embeddings which used as input. To evaluate the impact of the pre-trained word embeddings, randomly initialized word embeddings have been used separately.
\subsection{Parameters}
In this work, the dimension of character-level embeddings is set to 100 and that of the pre-trained word embeddings to 200. Different values for these parameters have been experimented and we set the values which reported the best performance for our experiments. Table \ref{parameters} shows the parameters and their values used in our experiments.
\begin{table}[!htb]
\renewcommand{\arraystretch}{1.2}
\begin{center}
\begin{tabular} {l|c}  \hline
\textbf{Parameter} & \textbf{Value}\\ \hline
 No. of epochs & 15          \\
 Dropout & 0.5         \\
 Batch size & 20           \\
 Learning method & Adam        \\
 Learning decay & 0.9        \\
 No. of LSTM units for character embeddings & 100 \\
 No. of LSTM units for word embeddings & 300 \\
 No. of distinct words in NCBI dataset & 8147 \\
 No. of distinct words in BC5CDR dataset & 13427 \\
 No. of distinct characters in NCBI dataset & 84 \\
 No. of distinct characters in BC5CDR dataset & 80 \\\hline
\end{tabular}
\end{center}
\caption{\label{parameters}Parameters and their values used for training the proposed model}
\end{table}
\section{Results and Discussion}
NCBI and BC5CDR datasets have been used to evaluate the proposed model. SR schemes namely IOB2 and IOBES have been used to train the model as these two schemes reported high performance among other schemes for BioNER as discussed in \cite{7732182}.\\
\\Results shown in Table \ref{resultvar} illustrate that the character embeddings significantly increase the performance for both datasets. Decoding the output vectors using CRF with global scores improves the performance rather than using local scores. The pre-trained word embedding results in better performance than randomly initialized word embeddings. The reason is that these pre-trained embeddings are trained over a collection of huge texts including texts from biomedical domain as well as generic domain. In both datastes, the best results are reported using character embeddings, pre-trained word embeddings, dictionary information and CRF with global scores. IOBES and IOB2 schemes do not show significant difference with NCBI dataset. However, using the same schemes with BC5CDR dataset shows a significant difference of 1.13 in the f1-measure.
\begin{table}[!h]
\renewcommand{\arraystretch}{1.30}
\begin{center}
\begin{tabular}{c|c|c|c|c|c|r}%{\textwidth}{@{\extracolsep{\fill}}ccccccc }  %{@{\extracolsep{\fill}} | c | c | c | r | }
\hline
\textbf{Dataset}&\textbf{SR scheme}&\textbf{V1}&\textbf{V2}&\textbf{V3 }&\textbf{V4}&\textbf{f1-measure}\\ \hline

\multirow{10}{*}{NCBI} & \multirow{5}{*}{IOB2}   &x          & x          & x          & x          & 71.16\%   \\
 &                          &x          & x          & x          & \checkmark & 80.46\%   \\
 &  		                &x          & x          & \checkmark & \checkmark & 83.13\%   \\
 &  		                &x          & \checkmark & \checkmark & \checkmark & 84.24\%   \\
 & 		                    &\checkmark & \checkmark & \checkmark & \checkmark & 85.19\%   \\ \cline{2-7}
 & \multirow{ 4}{*}{IOBES}  &x          & x          & x          & x          & 73.23\%   \\
 &                          &x          & x          & x          & \checkmark & 81.60\%   \\
 & 		                    &x          & x          & \checkmark & \checkmark & 83.44\%   \\
 & 		                    &x          & \checkmark & \checkmark & \checkmark & 84.40\%   \\
 & 		                    &\checkmark & \checkmark & \checkmark & \checkmark & \textbf{85.40}\%   \\ \hline
\multirow{10}{*}{BC5CDR} & \multirow{5}{*}{IOB2}   &x          & x          & x          & x          & 73.26\%   \\
 &                          &x          & x          & x          & \checkmark & 75.81\%   \\
 &  		                &x          & x          & \checkmark & \checkmark & 78.30\%   \\
 &  		                &x          & \checkmark & \checkmark & \checkmark & 78.36\%   \\
 & 		                    &\checkmark & \checkmark & \checkmark & \checkmark & 78.49\%   \\ \cline{2-7}
 & \multirow{ 4}{*}{IOBES}  &x          & x          & x          & x          & 73.64\%   \\
 &                          &x          & x          & x          & \checkmark & 76.98\%   \\
 & 		                    &x          & x          & \checkmark & \checkmark & 77.95\%   \\
 & 		                    &x          & \checkmark & \checkmark & \checkmark & 79.33\%   \\
 & 		                    &\checkmark & \checkmark & \checkmark & \checkmark & \textbf{79.62}\%   \\ \hline

		\end{tabular}
	\end{center}
	\caption{\label{resultvar}Results\vspace{0.3in} of the proposed model with following variations:\newline
             \textbf{V1}: (x) without dictionary information (\checkmark) with dictionary information\newline \textbf{V2}: (x) randomly initialized word embeddings (\checkmark) pre-trained word embeddings\newline \textbf{V3}: (x) using local score for decoding (\checkmark) using global score for decoding\newline \textbf{V4}: (x) without character embeddings (\checkmark) with character embeddings}
\end{table}
Table \ref{comparision1} and Table \ref{comparision2} give the comparisons between our model and the state-of-the-art models for NCBI and BC5CDR datasets respectively. For NCBI dataset, the performance of our model outperforms the state-of-the-art works.\\
\begin{table}[!h]
% Comparison for other results
\renewcommand{\arraystretch}{1.35}
\begin{center}
\begin{tabular} {l|c}  \hline
\textbf{Model}                                               & \textbf{f1-measure}\\ \hline
 ANN (BiLSTM + CNN)  \cite{sahu-anand:2016:P16-1}            & 79.13\%          \\
 Pairwise learning \cite{leamanDnorm}                        & 80.90\%          \\
 ANN (ReLU + Softmax activiation) \cite{Crichton2017}        & 80.74\%          \\
 TaggerOne (Semi-Markov model) \cite{Leaman09062016}         & 82.90\%          \\
 BANNER (CRF) \cite{leaman2008banner}                        & 81.80\%          \\
 CNN (Dictionary + Postprocessing) \cite{zhao2017disease}   & 85.17\%          \\  \hline
 \textbf{Our model}  (Best result)                                                     & \textbf{85.40}\% \\ \hline
\end{tabular}
\end{center}
\caption{\label{comparision1}Comparison of our model with related work on NCBI dataset}
\end{table}

\begin{table}[!h]
% Comparison for other results
\renewcommand{\arraystretch}{1.35}
\begin{center}
\begin{tabular}{l | c}  \hline
\textbf{Model}                                                             & \textbf{f1-measure}  \\ \hline
 LSTM \cite{bc5scdr7650}  & 76.50\% \\
 Ensemble (RNN + CRF) \cite{doi:10.1093/database/baw140}                   & 78.04\%            \\
  CD-REST (CRF + External resource) \cite{doi:10.1093/database/baw036}      & 84.43\%            \\
 CRF (dictionary Information)\cite{lee2015enhanced}                    & 86.46\% \\
 Ensemble(SVM+CRF) \cite{doi:10.1093/database/baw077}                      & 86.93\%            \\
 CNN (Dictionary + Postprocessing) \cite{zhao2017disease}                  & 87.83\%            \\\hline
 \textbf{Our model}   (Best result)                                                             & \textbf{79.62}\%   \\
 \hline
\end{tabular}
\end{center}
\caption{\label{comparision2}Comparison of our model with related work on BC5CDR dataset}
\end{table}

\section{Conclusion}
In this paper, an ANN-based model for Disease-NER is presented. Instead of hand engineering the features for tokens, character-level embeddings are used to represent orthographical features of tokens in addition to using word embeddings and dictionary information. Two different SR schemes namely IOB2 and IOBES are used for annotating the corpus. Results show that using character embeddings, pre-trained word embeddings, dictionary information and CRF with global scores improves the performance of BioNER. In addition, IOBES scheme outperforms IOB2 scheme.

\bibliographystyle{unsrt}
\bibliography{pre}
\end{document}